\documentclass[times,twocolumn,final,authoryear]{elsarticle}

\usepackage{ycviu}
\usepackage{framed,multirow}

\usepackage{amssymb}
\usepackage{latexsym}

\usepackage{url}
\usepackage[hidelinks]{hyperref}
\usepackage{xcolor}
\definecolor{newcolor}{rgb}{.8,.349,.1}

\usepackage{xspace}
\usepackage{times}
\usepackage{epsfig}
\usepackage{graphicx}
\usepackage{amsmath}
\usepackage{amssymb}
\usepackage{booktabs}
\usepackage{siunitx}
\usepackage{subfigure}
\usepackage{float}
\usepackage{algorithm}
\usepackage{algpseudocode}

\newcommand{\ie}{\textit{i}.\textit{e}.\@\xspace}
\newcommand{\eg}{\textit{e}.\textit{g}.\@\xspace}
\newcommand{\etc}{\textit{etc}.\@\xspace}

\algnewcommand\algorithmicforeach{\textbf{for each}}
\algdef{S}[FOR]{ForEach}[1]{\algorithmicforeach\ #1\ \algorithmicdo}

\journal{Computer Vision and Image Understanding}

\begin{document}

\begin{frontmatter}

\title{Anchor Pruning for Object Detection}

\author{Maxim \snm{Bonnaerens}\corref{cor1}} 
\cortext[cor1]{Corresponding author.\\
This work has been accepted to Elsevier - Journal of Computer Vision and Image Understanding for publication. It is available at \url{https://doi.org/10.1016/j.cviu.2022.103445}}
\ead{maxim.bonnaerens@ugent.be}
\author{Matthias \snm{Freiberger}}
\author{Joni \snm{Dambre}}

\address{IDLab-AIRO, Ghent University - imec, Gent, Oost Vlaanderen, Belgium}

\received{}
\finalform{}
\accepted{}
\availableonline{}
\communicated{}

\begin{abstract}
This paper proposes anchor pruning for object detection in one-stage anchor-based detectors.
While pruning techniques are widely used to reduce the computational cost of convolutional neural networks, they tend to focus on optimizing the  backbone networks where often most computations are. In this work we demonstrate an additional pruning technique, specifically for object detection: anchor pruning. 
With more efficient backbone networks and a growing trend of deploying object detectors on embedded systems where post-processing steps such as non-maximum suppression can be a bottleneck, the impact of the anchors used in the detection head is becoming increasingly more important.
In this work, we show that many anchors in the object detection head can be removed without any loss in accuracy. With additional retraining, anchor pruning can even lead to improved accuracy. Extensive experiments on SSD and MS COCO show that the detection head can be made up to 44\% more efficient while simultaneously increasing accuracy. Further experiments on RetinaNet and PASCAL VOC show the general effectiveness of our approach. We also introduce `overanchorized' models that can be used together with anchor pruning to eliminate hyperparameters related to the initial shape of anchors. Code and models are available at \url{https://github.com/Mxbonn/anchor_pruning}.
\end{abstract}


\end{frontmatter}


Object detection is one of the most widely studied tasks in computer vision. It requires not only predicting the class but also the exact location of each object in an image. Object detection models are mainly categorized into two categories based on whether detection happens in one or two stages. Two-stage networks (\cite{cai2018cascade,girshick2015fast,ren2015faster}), first run a regional proposal network (RPN) that produces class-agnostic candidate object locations. These proposals are then used as input for a second stage where the proposed object regions are classified and the location of the object is refined. One-stage networks (\cite{lin2017retinanet, liu2016ssd, redmon2017yolo9000, sermanet2013overfeat, tan2020efficientdet}) use a single convolutional neural network with a predefined set of default anchors (\ie an anchor associates a predefined size and aspect ratio to each pixel at a feature map) of which the classes and location offsets are directly predicted.
While two-stage networks are generally the most accurate, one-stage detectors tend to be more resource-efficient and are often the only choice to run in real-time on resource-constrained devices, such as mobile platforms (\cite{huang2017speed}). 

Many studies aimed at creating more efficient models for object detection have been carried out in recent years. Most have achieved it through slimmer backbone networks (\eg MobileNet in SSDLite (\cite{sandler2018mobilenetv2})) or compression techniques such as weight pruning and quantization (\cite{deng2020model, liu2018rethinking}). And while there has been some work on anchor-free detectors (\cite{law2018cornernet, tian2019fcos}), most mainstream single-shot detectors still use the same anchor-based approach (\cite{lin2017retinanet, redmon2018yolov3, tan2020efficientdet}).
As computational complexity in the backbone network keeps shrinking, the importance of an efficient object detection head, in which anchors are used to predict potential objects, increases.
The influence of these anchors on the computational cost of the object detector is twofold. Firstly, the number of anchors an object detector uses directly influences the size of the convolutional layers in the head of the detection model. Secondly, all bounding boxes produced by those anchors need to go through a non-maximum suppression (NMS) post-processing step to determine the final detected objects.
Previous studies on the speed/accuracy trade-off for object detectors have shown that for small models, \ie the ones typically used in an embedded context, ``NMS can take up the bulk of the running time" (\cite{huang2017speed}). While the exact running time of the post processing step depends strongly on the used hardware, systematic assessments (\cite{verucchi2020systematic, cai2019maxpoolnms}) have shown that the number of detections coming out of an object detection network has a non-negligible influence on the end-to-end inference time.
Whereas traditional backbone pruning techniques only influence part of the model inference time, anchor pruning can reduce the running time of both the model inference and the post-processing step.

In this paper we introduce a novel anchor pruning technique and evaluate its effect on the accuracy and efficiency of one-stage anchor-based object detection models. The main contributions of this work include:

(1) We propose an anchor pruning search algorithm that determines which key anchors to remove and identifies several anchor configurations on the accuracy/resources Pareto front, allowing model developers to trade accuracy for performance when resources are constrained.

(2) We show that the accuracy of the proposed pruned models can be further improved through fine-tuning or retraining from scratch. By removing well-chosen anchors and retraining, it is possible to achieve both better efficiency and accuracy compared to the unpruned baseline model.

(3) We introduce the concept of an `overanchorized' model. By starting our proposed anchor pruning approach on an object detector that has many closely overlapping anchors, we show that we can avoid tuning hyperparameters related to the initial shape of the anchors.

(4) We extensively benchmark our approach on the SSD and RetinaNet object detectors on both the MS COCO and PASCAL VOC datasets. The results strongly demonstrate that pruning anchors and retraining can reduce the resource costs of an object detector significantly and that our method generalizes to different anchor-based single shot detectors.

\section{Related Work}
\label{sec:related_work}
\subsection{Model scaling}
Many real-world object detection applications run on systems where model size and inference speed are highly constrained. The object detection networks used in such systems are not selected based on absolute state-of-the-art accuracy but on the best possible performance given these constraints. To this end, many model scaling techniques are applied to scale a model towards a certain size or speed, where the size is determined by the number of parameters in the network and the speed by the number of floating point operations. In object detection, it is common to downscale the model by changing the backbone network (\eg from VGG (\cite{simonyan2014very}) to MobileNet (\cite{sandler2018mobilenetv2})), by reducing the input resolution (\eg from $512 \times 512$ to $300 \times 300$ in SSD (\cite{liu2016ssd})), by reducing the number of layers (\eg from 50 to 18 in ResNet (\cite{he2016deep})) or a combination of these (\cite{tan2020efficientdet}). 
To further trim the computational cost in those models, techniques such as pruning (\cite{liu2018rethinking}) and quantization (\cite{zhou2017incremental}) are often used, possibly in combination with knowledge distillation (\cite{hinton2015distilling}) to recover any lost accuracy.
Earlier pruning techniques focused on removing individual weights or connections, but in recent years, the trend has shifted towards more structural pruning at the level of filters or even entire layers. Compared to pruning many individual weights, structural pruning does not need specialized hardware or libraries to benefit from the sparsity.

However, none of the above scaling techniques are specific to object detection. We will show in Section \ref{sec:anchor_pruning} that our proposed method of pruning anchors is an effective way to downscale the computational complexity of the object detection head. It is orthogonal to the previously mentioned scaling methods, meaning that they can be combined to achieve better compression results.

\subsection{Anchor-based}
Anchor-based one-stage detectors are convolutional object detection models that consist in most cases of three parts. 
The \textit{backbone} network is an off-the-shelf convolutional network, trained originally for image classification, that extracts features for detecting objects. On top of this backbone, most object detection models have a middle part, the \textit{neck}, that adds more layers after the backbone to produce better adapted features. In SSD (\cite{liu2016ssd}) the neck consists of additional convolutional feature layers, while in more recent networks such as RetinaNet (\cite{lin2017retinanet}) and EfficientDet (\cite{tan2020efficientdet}) some form of Feature Pyramid Network (FPN) (\cite{lin2017feature}) is used, that has additional lateral connections with the backbone. The final part is the \textit{object detection head} which includes a classifier and regressor to predict classes and exact locations of the objects in the image. The detection head is applied to several feature layers which can come from both the backbone and the neck.

Anchor-based detectors associate some predefined \textit{anchors}, sometimes also named default bounding boxes or priors, to each feature layer to which the detection head is attached. These anchor-associated feature layers are what we will from now on refer to as feature maps. Usually, the anchors are defined in terms of size and aspect ratio. The classifier and regressor in the detection head output the class scores and the 4 offsets relative to the predefined anchor shape for each pixel in a feature map. 
While many papers claim that an increase in the number of anchors leads to an increase in accuracy (\cite{li2019dynamic, liu2016ssd}), we will show in Section \ref{sec:experiments} that this does not always hold and that too many anchors can lead to a decrease in accuracy.

\subsection{Anchor-free}
Some one-stage detectors use a different approach. The first version of YOLO predicted the coordinates of bounding boxes directly using fully connected layers (\cite{redmon2016you}), the latest versions have however also switched to using anchors. More recently proposed anchor-free detectors have deliberately moved away from using anchors. CornerNet (\cite{law2018cornernet}) for example, detects corners of an object and groups multiple predictions together to produce a final bounding box. One of the reasons of existence for FCOS (\cite{tian2019fcos}), another anchor-free object detector, is that anchor shapes need to be carefully tuned.
While anchor-free methods are growing in popularity and show promising results, their alternatives to anchors often mean that the detections made in the object detection head are fixed and cannot be downscaled to trade accuracy for performance.
We believe that anchor-based solutions are still relevant as they are widely used and as demonstrated in EfficientDet (\cite{tan2020efficientdet}), are able to achieve state-of-the art results in both accuracy and efficiency.

\subsection{Anchor shape optimization}
One of the reasons to use an anchor-free approach is that the number of anchors and their predefined shapes can be seen as hyperparameters that need to be carefully tuned.
As stated earlier, anchors are usually defined in terms of size (or scale) and aspect ratio.
The optimal aspect ratios depend strongly on the domain in which the object detector is applied. For face recognition square aspect ratios will be important (\cite{zhang2017s3fd}) while for text detection such as in license plate recognition wider aspect ratios like $1:5$ need to be included (\cite{liao2018textboxes++}).

The optimal scales for the anchors in each layer, not only depend on the domain (\ie the distribution of the size of the objects) but also on the receptive field and the stride of the associated feature map. The theoretical receptive field indicates which pixels in the input image affect the value of a pixel in the feature map. However as shown in (\cite{luo2016understanding}), the effective input region that has a non-negligible impact on this feature map value is only a fraction of the theoretical receptive field. The stride of the feature map determines the interval at which anchors are placed. For example, in SSD the stride size of the first anchor-associated feature map is 8 pixels, indicating that the anchors produces a bounding box every 8 pixels on the input image.

To reduce the importance of the initial anchor shapes, 
RefineDet (\cite{zhang2018refinedet}) proposes a one-stage object detector with an additional module to refine anchor locations, alleviating the negative effects of suboptimal anchor shapes. Other approaches such as Guided Anchoring (\cite{wang2019guidedanchoring}) and MetaAnchor (\cite{yang2018metaanchor}) remove the need for hand-picked anchors all together by including additional components that generate the anchors. However, this requires extra resources and results in a larger model and longer inference time. 
More recent approaches dynamically learn the anchors during training, which increases training time but has no influence on the model size and inference time. FreeAnchor (\cite{zhang2019freeanchor}) and MAL (\cite{ke2020multiple}) start from a bag of candidate anchors and learn which ones are optimal, Anchor Box Optimization (\cite{zhong2020anchor}) adapts the loss and changes the anchor shapes during back-propagation.
These approaches optimize the anchor shape initialization but still leave the number of anchors as a hyperparameter. While these  approaches benefit from anchor selection during training, this also means that any modification to the numbers of anchors requires retraining the network, making it infeasible to use such methods to efficiently explore accuracy/performance trade-offs. In comparison our proposed pruning method can obtain an accuracy without any need for re-training or fine-tuning. Only when a pruned configuration is selected does our method require fine-tuning or retraining. 

Our anchor pruning approach is compatible with these previous anchor shape optimization methods as it does not change the shapes of the initial anchors but rather removes the redundant or least important anchors. Models that have optimized anchor shapes can be used as a baseline from which to start anchor pruning.

\section{Anchor Pruning}
\label{sec:anchor_pruning}
In this section, we propose an efficient way to explore different anchor configurations through pruning.
We first demonstrate why pruning anchors is a sensible strategy. Next, we present a search algorithm to determine which anchors to remove for different resource constraints. The result is a sequence of anchor configurations that form a Pareto front in the accuracy/performance design space. This allows model designers to trade-off between accuracy and performance when selecting their final model. 

\subsection{Redundant Anchors}
In a one-stage anchor-based object detector, the total number of predicted bounding boxes per image is $ N = \sum_{i=1}^{k} N_i = \sum_{i=1}^{k} (A_i \times H_i \times W_i)$, where $H_i$ and $W_i$ are the size of the feature maps, and $A_i$ is the number of anchors, for $k$ different feature map layers. In some object detection architectures $N$ also depends on the number of classes $C$ but since most recent models separate class-agnostic bounding box regression from the object classification, we do not include it here. Our proposed technique can be trivially extended to handle class-specific anchors.

When $\forall i,j \in 1,...,k: A_i = A_j$ and $H_{i+1} = H_i / 2$, $W_{i+1} = W_i / 2$, as is common in single stage detectors where the feature maps are often in a pyramid structure, then $N_1 \approx \frac{3 N}{4}$. This means that the majority of the predicted bounding boxes come from the anchors associated with the first feature map.

Table \ref{tab:example_1} shows the relation between the number of anchors per feature map layer and the total number of bounding boxes as well as the  accuracy on the COCO dataset for different SSD300 models. The first row shows the results when using the 6 anchors per layer as defined in the scales and aspect ratios section of the original SSD paper (\cite{liu2016ssd}). The second row shows the results for the final SSD300 version reported in the experimental results section of the original paper where the number of anchors is reduced from 6 to 4 in the first and last two layers. While the original paper does not motivate why some layers have only 4 instead of 6 anchors, it can be seen from the results in the table that the choice likely comes from the observation that 6 anchors in the first layer offer less accuracy at a higher computational cost. The last row in Table \ref{tab:example_1} shows that further reducing the number of anchors in that first layer could have improved accuracy and efficiency even more. As we argued earlier, the optimal number of anchors is a hyperparameter that requires careful tuning. This example also indicates that more anchors do not always mean better accuracy, as will be further shown in section \ref{sec:experiments}.

\begin{table}[h]
\centering
\caption{Relation between the number of anchors per layer $A_i$ and the total number of bounding boxes $N$ in SSD300.
The number of anchors $A_1$ in the first feature map layer have the largest influence on $N$. It can be seen that reducing this number has a positive influence on both the mean average precision (mAP) and the number of bounding boxes. \newline \textbf{*} Marks the original configuration of SSD300 (\cite{liu2016ssd}).}
\begin{tabular}{c c c} \toprule
$\{A_{1},A_{2},A_{3},A_{4},A_{5},A_{6}\}$ & $N$ & COCO mAP \\ \midrule
$\{6,6,6,6,6,6\}$ & 11640 & 25.6 \\
$\{4,6,6,6,4,4\}$\textsuperscript{\textbf{*}} & 8732 & 25.7 \\
$\{2,6,6,6,4,4\}$ & 5844 & 25.8 \\ \bottomrule
\end{tabular}
\label{tab:example_1}
\end{table}

\begin{figure}[ht]
    \centering
    \includegraphics[width=\linewidth]{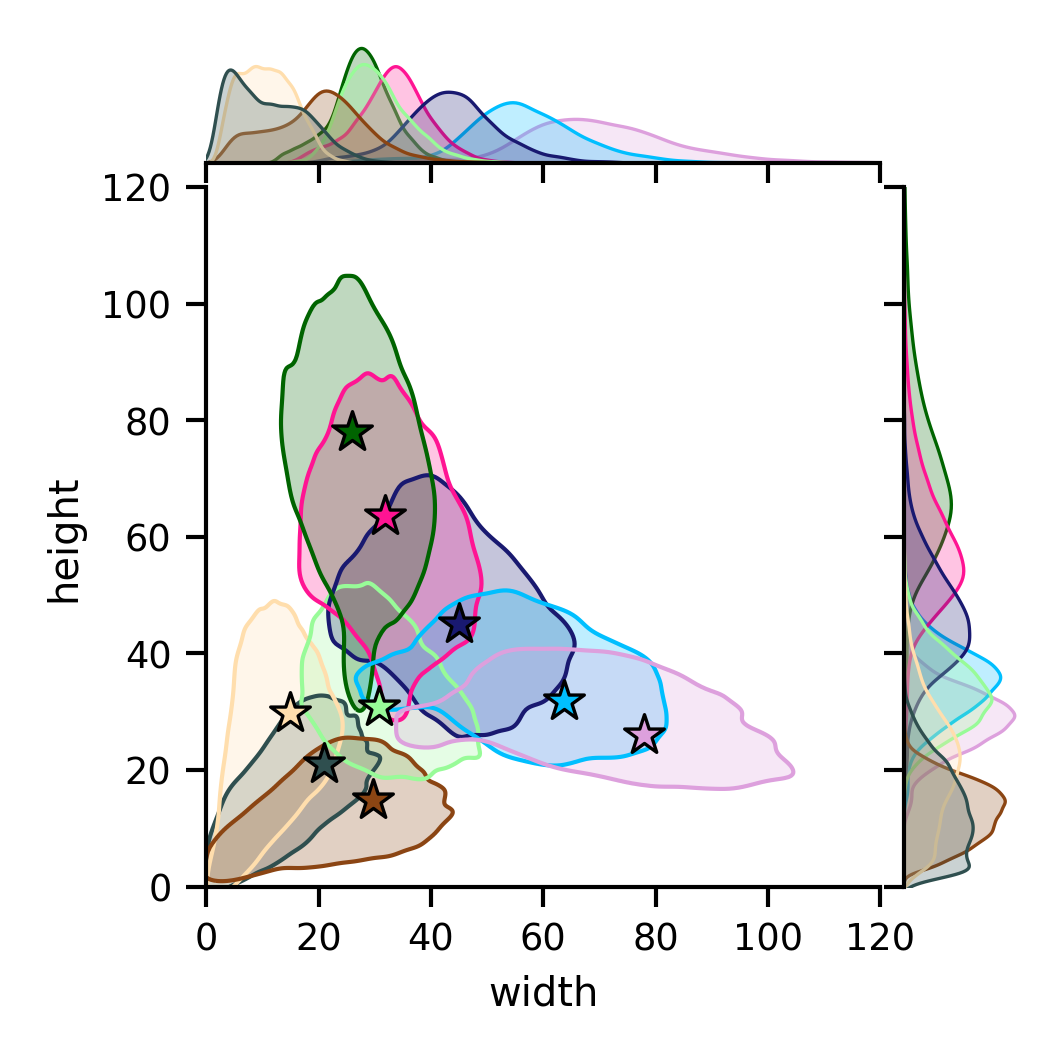}
    \caption{The distribution of the shape of bounding boxes for each anchor in the first two layers of SSD300 on MS COCO. The markers correspond to the default shape of each anchor, and the matching colored region to the final bounding boxes produced by that anchor. Plotted on the side are the marginal distributions of the width and height.}
    \label{fig:anchor_distribution}
\end{figure}

Analysis of the  bounding boxes predicted by each anchor, shows that different anchors can predict similar bounding box shapes. For example, an object with a ground truth bounding box of aspect ratio $1:1.5$ could be predicted by a $1:1$ anchor and a $1:2$ anchor.
Figure \ref{fig:anchor_distribution}, shows the distribution of the predicted bounding boxes for each anchor in the first two layers of SSD300 on the MS COCO dataset. It can be observed that certain anchors produce bounding boxes that are almost completely covered by predictions of neighboring anchors, which intuitively explains why it is possible to drop certain anchors without loss of performance.

\subsection{Optimal Anchor Search}
Unlike methods such as magnitude-based pruning of convolutional filters (\cite{li2016pruning}), we not only have to decide how much to prune but also what to prune. Anchors that produce bounding boxes that are largely covered by neighboring anchors are more optimal to prune than those producing unique shapes.
The number of possible anchor subsets that result from pruning a model is equal to $|\mathcal{P}(\mathcal{A})| = 2^{|\mathcal{A}|}$ where $\mathcal{A}$ is the set of all anchors in the unpruned model and $\mathcal{P}(\mathcal{A})$ stands for the power set of $\mathcal{A}$. Models such as SSD and RetinaNet have 30 and 45 anchors respectively, which makes it infeasible to evaluate all combinations.
We therefore introduce a greedy search algorithm to efficiently explore the search space. This procedure is summarized in Algorithm \ref{alg:anchor_pruning_search}. 

\begin{algorithm}
\caption{Anchor Pruning Search}    
\hspace*{\algorithmicindent} \textbf{Input:} Fully trained model $M$ with anchors $\mathcal{A}$\\
\hspace*{\algorithmicindent} \textbf{Output:} Pareto Frontier $P$
\begin{algorithmic}[1]
\State Set $S \gets \{\mathcal{A}\}$ and $P \gets \{\mathcal{A}\}$
\While{$S \ne \emptyset$}
\State Select a configuration $\mathcal{C}$ from $S$ to explore
\ForEach{anchor $a_i \in \mathcal{C}$}
\State $\mathcal{C}_i = \mathcal{C} \setminus a_i$
\State $\texttt{accuracy}_{\mathcal{C}_i}$ = Accuracy of $M$, keeping only \newline 
\hspace*{2.7em} predictions produced by anchors in $\mathcal{C}_i$
\State $\texttt{cost}_{\mathcal{C}_i}$ = Resource cost of $M$ with anchors in $\mathcal{C}_i$
\State Compare $\texttt{accuracy}_{\mathcal{C}_i}$ and $\texttt{cost}_{\mathcal{C}_i}$ to \newline 
\hspace*{2.7em} $\texttt{accuracy}_{\mathcal{C}_j}$ and $\texttt{cost}_{\mathcal{C}_j}$ from all $\mathcal{C}_j \in P$
\If{$\mathcal{C}_i$ is Pareto Optimal in $P$}
\State Add $\mathcal{C}_i$ to $P$ and $S$ 
\EndIf
\State Remove any $\mathcal{C}_i$ that is no longer optimal from $P$
\EndFor
\EndWhile
\end{algorithmic}
\label{alg:anchor_pruning_search}
\end{algorithm}

Our anchor pruning algorithm starts from a fully (pre-)trained model $M$ with anchor configuration $\mathcal{C} = \mathcal{A}$. 
The algorithm iterates over an adapting set of unexplored anchor configurations $S$ and constructs a set of Pareto-efficient anchor configurations $P$.
Starting from $S = \{\mathcal{A}\}$ and $P = \{\mathcal{A}\}$, we take an unexplored configuration $\mathcal{C}$ ($\mathcal{C} \subseteq \mathcal{A}$) from $S$ at each iteration until $S = \emptyset$. 
From anchor configuration $\mathcal{C}$ we evaluate configurations $\mathcal{C}_1, ..., \mathcal{C}_n$, where $\mathcal{C}_i = \mathcal{C} \setminus a_i$, with anchor $a_i \in \mathcal{C}$.

Evaluating a configuration $\mathcal{C}$ means evaluating the accuracy of the model $M$ when only using the predictions made by anchors $a_i \in \mathcal{C}$.
This can be done in an efficient way by initially storing all predicted bounding boxes from $M$, \ie before any are discarded by post-processing steps such as NMS, and to re-evaluate by only keeping the bounding boxes produced by anchors $a_i$ in the configuration $\mathcal{C}$. This avoids the costly process of running all validation inputs through an adapted model for each new configuration.
The evaluation of the accuracy metric should be done on a validation set to avoid overfitting the anchor configurations on the test set. 
As stated before, we are not just interested in the highest accuracy but in the accuracy/resources trade-off. This requires evaluating an accuracy metric (\eg mean average precision (mAP), F-score, \etc) and a resource metric (\eg inference speed, FLOPs, model size, number of predicted bounding boxes, \etc).
Resource metrics that can be calculated such as FLOPs should be preferred above metrics that need to be experimentally measured such as inference speed. 

A configuration is Pareto-optimal or Pareto-efficient when no other node that uses fewer or equal resources achieves higher accuracy.
At any point during the search phase, $P$ stores the Pareto-optimal configurations that have been evaluated thus far.
To achieve this, the accuracy and resource cost of $\mathcal{C}_i$ are compared to all $\mathcal{C}_j \in P$, if $\mathcal{C}_i$ is Pareto optimal in $P$ we add $\mathcal{C}_i$ to $P$ and $S$, and any $\mathcal{C}_j$ that is no longer Pareto optimal is removed from $P$.
To speedup the running time of the search algorithm, one could also add an accuracy threshold $\theta$ such that $\mathcal{C}_i$ is only added to $P$ and $S$ if it is Pareto Optimal in $P$ and $\texttt{accuracy}_{\mathcal{C}_i} \ge \theta$. This is often useful in practice when we are only interested in accuracy/resources trade-offs as long as the accuracy remains acceptable.
Once $S$ is empty, $P$ contains all Pareto-efficient anchor configurations. 

Because the search is greedy, $P$ might contain configurations that are suboptimal solutions as the algorithm may miss anchor combinations that would arise from further pruning discarded configurations. However, as we will show in Section \ref{sec:experiments}, our method significantly outperforms random pruning and the default pruning approach from RetinaNet.

\section{Experiments} 
\label{sec:experiments}
\subsection{Setup}
\label{sec:experiments_setup}
Our proposed anchor pruning is rather general and can be applied to any anchor-based one-stage object detector.
We demonstrate our results primarily on SSD300 (\cite{liu2016ssd}), one of the most used and influential one-stage object detectors.
SSD uses a VGG16 backbone, a pyramid of convolutional feature maps in the neck, and a head with a $3 \times 3 \times (A_i \times (Classes + 4))$ convolution on top of each of the 6 feature maps. Each feature map has anchors of a fixed scale and with aspect ratios of $\{1,2,\frac{1}{2},3,\frac{1}{3}\}$ and an additional anchor with aspect ratio $1$ but a larger scale (which we will refer to as $1^{+}$), resulting in 6 anchors per layer. However for the first and the two last feature maps, the anchors with aspect ratios $3$ and $\frac{1}{3}$ are removed, resulting in a total of $30$ anchors.

While SSD is no longer a state-of-the-art object detection model, as many alternatives that are both more accurate and more efficient have been developed since its publication, the fundamental structure of the Single-Shot-Detector remains used in most recent object detectors.

Table \ref{tab:flops_detectors} shows the FLOPs (multiply-adds) of different object detection models along with how many of those FLOPs are in the detection head.
It can be seen that the relative importance of the head has increased as backbones got more efficient in more recent object detectors. 
Given the growing focus on resource-efficient models, we expect this trend to continue. We will show in Section \ref{sec:retinanet} that our method generalizes well to these more recent anchor-based one-staged object detectors.

As stated in the introduction, the running time of an object detection model in an embedded context is often dominated by the running time of the post processing steps, which is directly related to the number of bounding boxes produced by the network. In systems where this post-processing bottleneck is not present, the running time will be directly related to the FLOPs of the model, and as our approach is backbone independent we only report on those FLOPs in the head. 
In the remainder of this work we report the resource cost as FLOPs in the head of the network or as the number of bounding boxes it produces. 
Compared to reporting latency which is very implementation and hardware dependent, we believe that our reported metrics allow model designers to better estimate the accuracy/resource trade-offs that are possible for their platform.

\begin{table}[h]
\centering
\caption{The number of FLOPs for different one-stage object detectors, along with the percentage of the FLOPs that take place in the head of the network. FLOPs are calculated for an input resolution of $300 \times 300$.}
\begin{tabular}{l c c} \toprule
Model & FLOPs & FLOPs(\%) head \\ \midrule
SSD & 34.4B & 12.3\% \\
RetinaNet & 21.7B & 57.6\% \\
MobileNetV3-Small-SSDLite & 0.2B & 37.5\% \\ \bottomrule
\end{tabular}
\label{tab:flops_detectors}
\end{table} 

\begin{table*}[th]
\centering
\caption{Results on COCO \texttt{test-dev2017} from the evaluation server for different anchor configurations of SSD. The models are ordered in decreasing number of bounding boxes and FLOPs. AP\textsubscript{.5:.95} is the COCO mAP, AR the average precision and subscripts S/M/L correspond to small, medium and large objects. The superscripts \textsuperscript{2,3,4,5} refer to the subsection of Section \ref{sec:experiments} where the details of the relevant experiments can be found.}
\begin{tabular}{l|ccc|ccc|ccc|cc} \toprule
Model (SSD variant) & AP\textsubscript{.5:.95} & AP\textsubscript{50} & AP\textsubscript{75} & AP\textsubscript{S} & AP\textsubscript{M} & AP\textsubscript{L} & AR\textsubscript{S} & AR\textsubscript{M} & AR\textsubscript{L} &\begin{tabular}[c]{@{}c@{}}FLOPs\\\textit{head}\end{tabular} & BBoxes \\ \midrule
SSD Baseline & 25.7 & 44.0 & 26.6 & 7.1 & 27.1 & 41.6 & 11.2 & 39.9 & 57.6 & 4231M & 8732 \\
SSD $\{1,2,\frac{1}{2},1^+\}^4$ & 25.5 & 43.7 & 26.2 & 7.3 & 26.6 & 40.8 & 11.6 & 39.6 & 56.5 & 3577M & 7760 \\
SSD $\{1,1^+\}^{4,5}$ & 25.0 & 43.3 & 25.5 & 7.2 & 25.6 & 39.6 & 12.7 & 39.0 & 56.1 & 1788M & 3880 \\
\midrule
Configuration-A \textit{pruned}$^{2,3}$ & 25.6 & 43.9 & 26.5 & 7.1 & 27.1 & 41.2 & 11.2 & 39.8 & 57.2 & 3607M & 7814 \\
Configuration-A \textit{fine-tuned}$^3$ & 25.7 & 44.0 & 26.6 & 7.2 & 27.2 & 41.3 & 11.3 & 39.7 & 57.3 & 3607M & 7814\\
Configuration-A \textit{retrained}$^3$ & 25.5 & 44.0 & 26.3 & 7.6 & 27.2 & 40.4 & 11.8 & 39.8 & 56.4 & 3607M & 7814\\
\midrule
Configuration-B \textit{retrained}$^{3,4}$ & 25.8 & 44.5 & 26.5 & 6.8 & 27.2 & 41.2 & 12.0 & 40.1 & 57.3 & 2476M & 4926 \\ 
Configuration-C \textit{retrained}$^{3,4,5}$ & 25.4 & 44.3 & 25.7 & 6.5 & 25.7 & 41.6 & 12.5 & 38.2 & 58.0 & 1628M & 3121 \\
Configuration-D \textit{retrained}$^3$ & 23.1 & 41.5 & 23.1 & 3.7 & 22.6 & 41.4 &7.3 & 35.4 & 58.0 & 774M & 1291 \\ 
\midrule
Pruned Layer-wise$^4$ & 25.0 & 43.6 & 25.5 & 7.5 & 26.5 & 38.1 & 13.0 & 39.3& 54.8 & 1788M & 3880 \\
\midrule
Overanchorized$^5$ & 25.8 & 43.5 & 27.0 & 6.6 & 28.4 & 42.1 & 9.7 & 40.7 & 59.1 & 6673M & 13584 \\
Pruned Overanchorized$^5$ & 25.3 & 44.0 & 25.7 & 6.5 & 25.6 & 42.1 & 12.3 & 37.4 & 58.8 & 1620M & 3080\\
\bottomrule
\end{tabular}
\label{tab:summarized_results}
\end{table*}

We evaluate pruning anchors of the SSD model on the MS COCO 2017 detection dataset \cite{lin2014microsoft}.
Before pruning, the SSD model is trained on \texttt{train2017} using SGD for 120 epochs with a learning rate of $10^{-3}$, which is decreased to $10^{-4}$ on epoch 80 and to $10^{-5}$ on epoch 110, with a weight decay of $5 \times 10^{-4}$ and a momentum of $0.9$. The input images are resized to $300 \times 300$ and all data augmentations as described in (\cite{liu2016ssd}) are used. The accuracy metric used for COCO is the mean average precision (mAP) evaluated at intersection-over-union (IoU) thresholds evenly distributed between 0.5 and 0.95, for the resource metric we use the number of FLOPs in the object detection head.
Our anchor pruning search, uses \texttt{val2017} to evaluate candidate anchor configurations.

\subsection{Pruning}
\begin{figure}[ht]
    \centering
    \includegraphics[width=\linewidth]{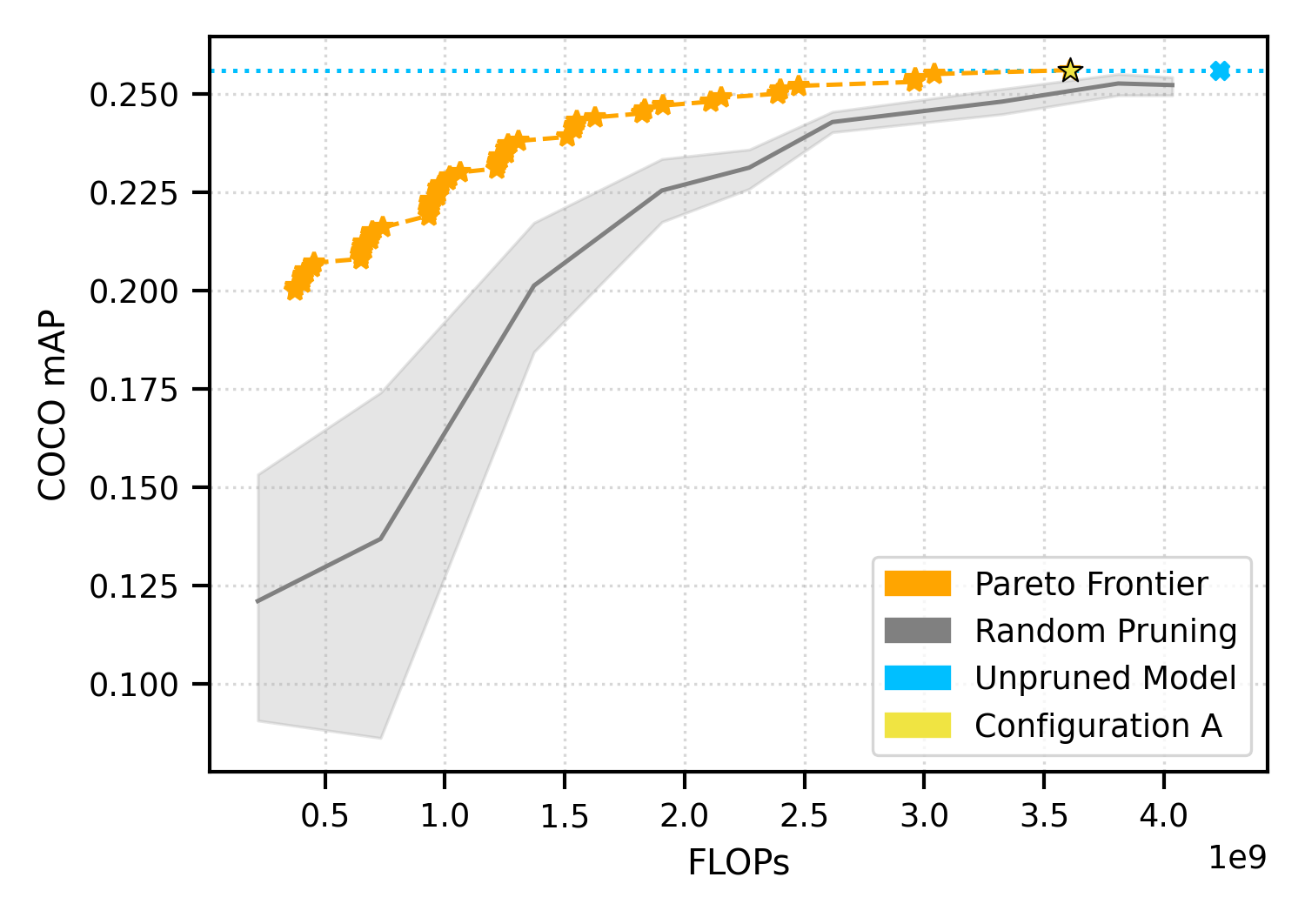}
    \caption{Accuracy and FLOPs trade-offs for different optimal anchor configurations found by our anchor pruning search. The Pareto frontier is highlighted in orange, the unpruned baseline in blue. Accuracy is measured on the COCO \texttt{val2017} dataset and FLOPs indicate the multiply-adds in the detection head.}
    \label{fig:pareto_front}
\end{figure}

Figure \ref{fig:pareto_front} shows the Pareto frontier of pruned anchor configurations as obtained by our search algorithm, representing points at which fewer FLOPs can only be achieved by sacrificing accuracy. We also highlight the original unpruned model and plot the results of configurations achieved when pruning anchors randomly. 
As can be seen in the figure, our method outperforms random pruning for all FLOPs, indicating our effectiveness in pruning anchors that are redundant or that lead to minimal accuracy loss. For pruned models that have a large reduction in FLOPs and therefore require more anchors to be pruned, selecting the right anchors can be the difference between a model with acceptable accuracy and a model that is no longer usable because the accuracy dropped too much.
Note that the wider gaps on the x-axis between anchor configurations in the Pareto frontier are due to pruned anchors that have a large influence on the number of FLOPs, such as anchors in the first layer of the head.

The most accurate configuration produced by our search algorithm, which we will refer to as Configuration-A \textit{pruned}, is able to reduce the number of FLOPs in the detection head by 15\% without losing any accuracy on the \texttt{val2017} dataset. Table \ref{tab:summarized_results} shows the accuracy in more detail on the \texttt{test-dev2017} dataset, indicating that the accuracy degraded slightly for large objects. This is not surprising as Configuration-A pruned 5 of the 8 largest anchors.

\subsection{Fine-tuning versus Retraining}
\begin{figure*}[ht]
    \centering
    \includegraphics[width=\textwidth]{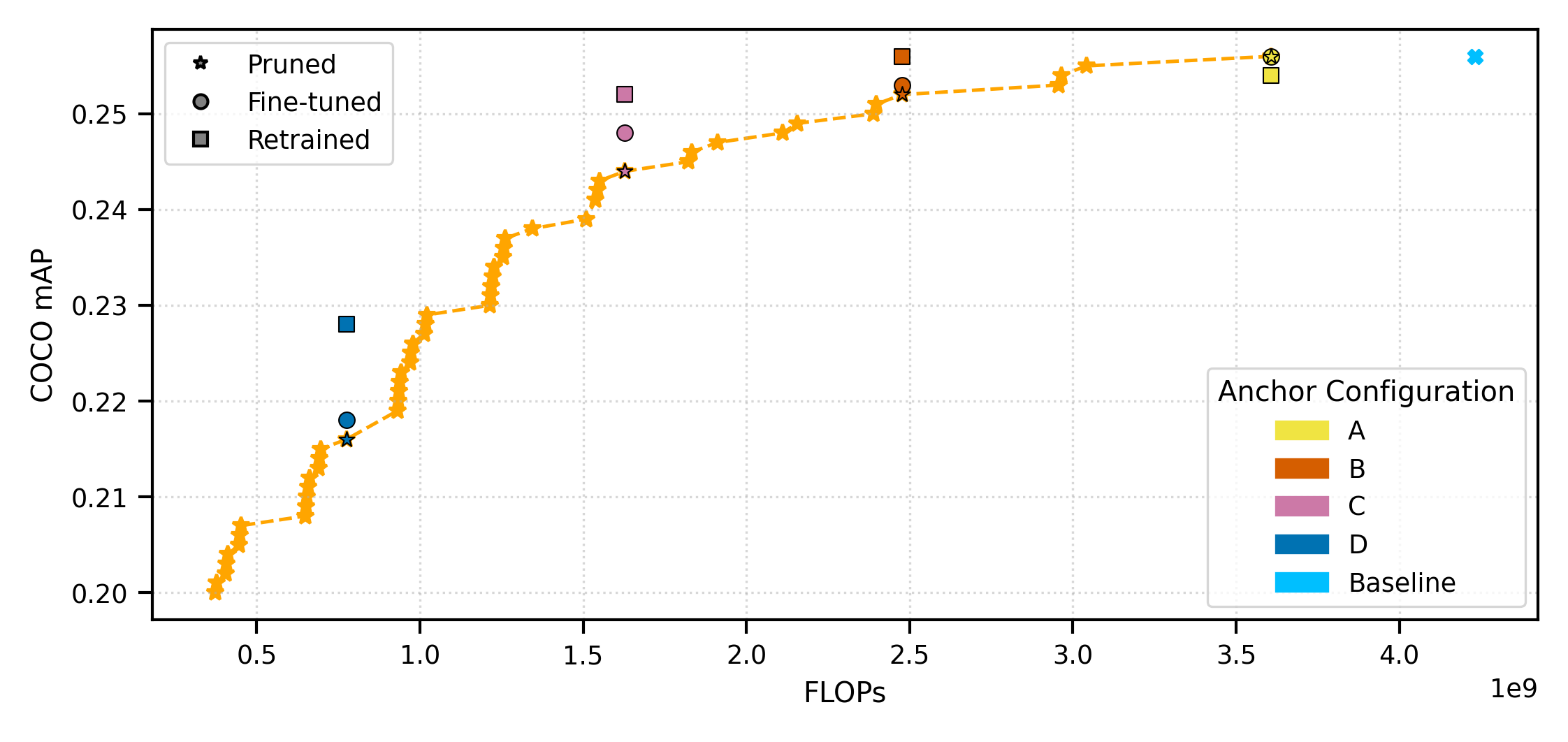}
    \caption{Accuracy of fine-tuning and retraining certain highlighted configurations for SSD on COCO \texttt{val2017}. Fine-tuning improves accuracy after pruning but in most cases, retraining achieves a much higher accuracy.}
    \label{fig:pareto_front_2}
\end{figure*}

Most pruning techniques fine-tune the model after pruning to recover part of the lost accuracy. In this section, we show the impact of fine-tuning after pruning anchors and compare it to retraining the model from scratch with the new anchor configuration.
When fine-tuning we train for 10 additional epochs with a learning rate of $10^{-5}$, for retraining we use the same settings as described in Section \ref{sec:experiments_setup}.
Figure \ref{fig:pareto_front_2} shows the effect of fine-tuning and retraining on four highlighted configurations. 
As illustrated in the figure, fine-tuning does recover some of the lost performance during pruning. However, for all nodes except Configuration-A, retraining the configuration from scratch achieves much higher accuracy. As more anchors get pruned, such as in Configuration-D, the difference between fine-tuning and retraining increases. The Pareto frontier produced by Algorithm \ref{alg:anchor_pruning_search} can be seen as a lower bound for the accuracy that can be achieved for those configurations after fine-tuning or retraining.

On the \texttt{val2017} dataset there are now multiple configurations that match the unpruned baseline accuracy: Configuration-A \textit{pruned}, Configuration-A \textit{fine-tuned} and Configuration-B \textit{retrained}. Further evaluation on the \texttt{test-dev2017} dataset, as shown in Table \ref{tab:summarized_results}, indicates that Configuration-A \textit{fine-tuned} is able to restore the lost performance as reported in the previous section and Configuration-B \textit{retrained} is able to \textbf{reduce the FLOPs and the number of bounding boxes by around 43\% while even slightly improving accuracy}.

\subsection{Layer-wise Pruning}
Experimenting with anchor configuration in the head of an object detector is not new, but it is normally done before training and in a symmetric way. Adding or removing anchor shapes is done across all layers and when the aspect ratio $x$ is added/removed, $\frac{1}{x}$ is also added/removed. The original SSD paper does mention two additional anchor settings compared to the baseline; one where aspect ratios $1$ and $\frac{1}{3}$ are removed, and one which is further reduced by removing aspect ratios $2, \frac{1}{2}$. Just like our retraining approach, these additional models are trained from scratched rather than created by pruning the larger baseline.
We compare these `default' settings to configurations in our Pareto frontier that have a comparable number of FLOPs. The SSD setting where aspect ratios $3$, $\frac{1}{3}$ are removed has a performance drop of 0.2\% on COCO. In the previous subsection, we already showed that we can prune much more than that while simultaneously improving accuracy. The SSD setting with only two square anchors per layer has a performance drop of 0.7\%. Configuration-C \textit{retrained} has a comparable number of FLOPs to this setting but only results in 0.3\% reduced accuracy. We also compare to a pruned model where the anchors are pruned layer by layer until 2 anchors per layer remain. While the layer-wise pruning configuration keeps other aspect ratios besides $1:1$, it does result in comparable accuracy (after retraining) to the last mentioned SSD setting, indicating the importance of pruning freely over all layers as we do in Algorithm  \ref{alg:anchor_pruning_search}. The detailed results can be found in Table \ref{tab:summarized_results}.

\subsection{Overanchorized Model}
\label{sec:overanchorized}
\begin{figure}[ht]
    \centering
    \includegraphics[width=.8\linewidth]{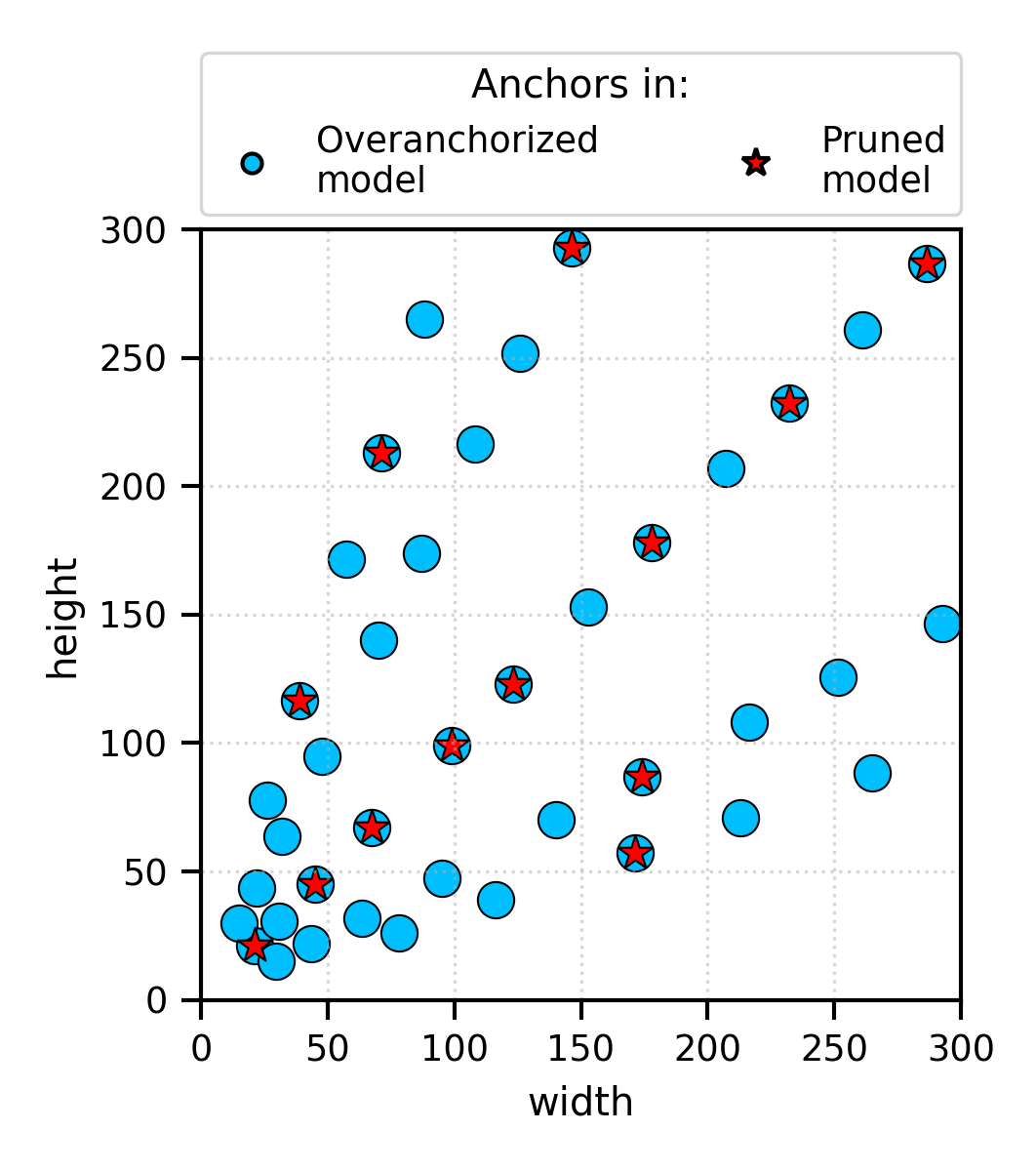}
    \caption{Anchor shapes for the `overanchorized' SSD model and the anchors that remain after pruning. The overanchorized model has 48 anchors that produce 13584 bounding boxes and achieves 25.8 mAP on COCO. The pruned version only has 14 anchors producing 3080 bounding boxes and achieves 25.4 mAP.}
    \label{fig:overanchorized}
\end{figure}

As explained in Section \ref{sec:related_work}, predefined anchors introduce many additional hyperparameters that require careful tuning.
We introduce an `overanchorized' model and define it as an object detection model that has more anchors than strictly needed. For applications where computational resources are not constrained, our previous experiments showed that while more anchors do not always improve accuracy, they also do not degrade the accuracy significantly.
For applications with constrained-resources we can prune from an overanchorized model to arrive at an optimized anchor configuration.
This also allows our algorithm to find a suitable set of anchors if no initial anchors with good performance are known.
As an experiment, we defined an overanchorized SSD model with 48 anchors (see Figure \ref{fig:overanchorized}) and pruned it to a model where the number FLOPs is below or equal to the SSD setting with two anchors per layer. Figure \ref{fig:overanchorized} plots the anchors selected by the pruning step. The detailed results in Table \ref{tab:summarized_results} show that pruning an overanchorized model achieves comparable accuracy to pruning the original baseline. This suggests that training an overanchorized object detection model followed by anchor pruning can eliminate extensive tuning of the anchor shapes, while simultaneously removing the number of anchors as a hyperparameter.

\subsection{RetinaNet}
\label{sec:retinanet}
As stated earlier, our approach is general as it can be applied to different one-stage anchor-based object detectors. In this subsection we demonstrate that our technique is also successful on more recent object detectors such as RetinaNet. 
The default configuration of RetinaNet uses 9 anchors in each layer by combining 3 scales and 3 aspect ratios $\{\frac{1}{2},1,2\}$.
As shown earlier in Table \ref{tab:flops_detectors}, unlike most object detectors, the majority of FLOPs in this model take place in the head. The explanation for this is that whereas the head in SSD directly does a $3 \times 3 \times (A_i \times (Classes + 4))$ convolution, RetinaNet first applies 4 additional $3 \times 3 \times 256$ convolutional layers. Another difference with SSD is that these prediction layers in the detection head are reused on all feature maps. To account for these differences, we apply the following changes to Algorithm \ref{alg:anchor_pruning_search}: Configuration $\mathcal{C}$ generates $\mathcal{C}_i$ by either removing a certain anchor configuration on each layer or by removing all anchors of a certain layer (as the prediction layers are shared, this practically means not applying the head on a certain feature map).
\begin{figure}[ht]
    \centering
    \includegraphics[width=\linewidth]{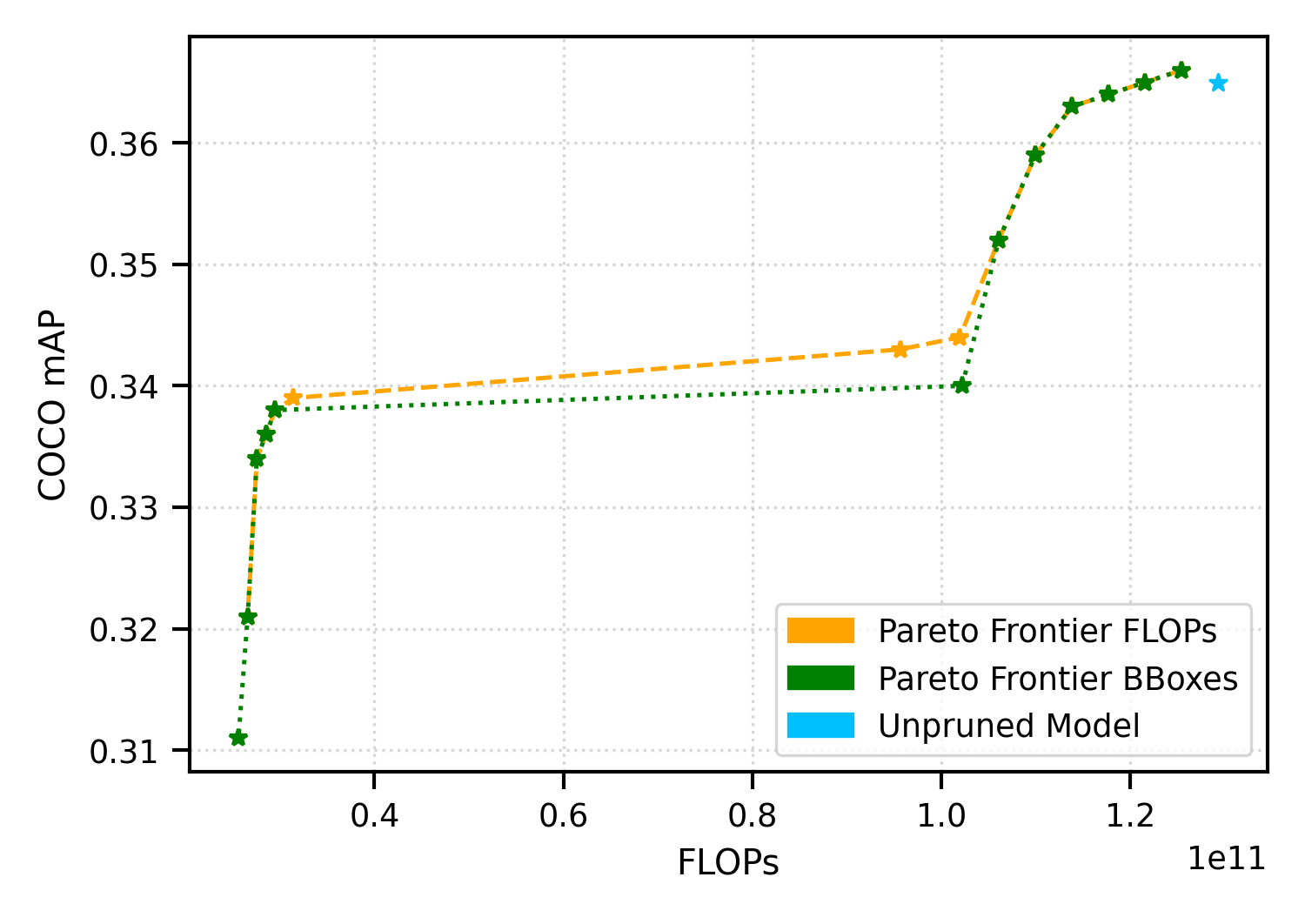}
    \caption{Accuracy and number of FLOPs for pruned anchor configurations of RetinaNet on COCO \texttt{val2017}. The Pareto frontiers are optimized for the number of FLOPs or the number of bounding boxes.}
    \label{fig:pareto_front_3}
\end{figure}

An important side effect of the additional convolutional layers in the RetinaNet head, is that the decision not to apply the head to a certain feature map has a large influence on the number of FLOPs but not necessarily on the number of bounding boxes. Because of this property, we use RetinaNet to show the difference between optimizing for the number of FLOPs or optimizing the number of bounding boxes.
When optimizing an object detector for a certain use-case, the used hardware will determine whether the computations in the convolutional layers or the post-processing steps on the bounding boxes are the most time consuming.
The resulting Pareto frontiers can be found in Figure \ref{fig:pareto_front_3}.
The large jump in FLOPs in both frontiers happens when the head is no longer applied to the first feature map. The difference is that in the FLOPs frontier the remaining layers still have 8 anchors left, while in the bounding boxes frontier there are only 5 anchors left. 

\begin{table}[ht]
\centering
\caption{Accuracy, FLOPs and inference time for RetinaNet variants on COCO \texttt{test-dev2017}. Our pruned model can reduce the FLOPs by 75\% compared to the baseline (s3a3) while being much more accurate than the smallest original RetinaNet head (s1a1). Inference times are measured on an Nvidia GTX 1080.}
\begin{tabular}{l|c|cc|c} \toprule
Model & AP\textsubscript{.5:.95} & \begin{tabular}[c]{@{}c@{}}FLOPs\\\textit{head}\end{tabular} & \begin{tabular}[c]{@{}c@{}}FLOPs\\\textit{total}\end{tabular} & inf time \\ \midrule
RetinaNet(s3a3) & 36.9 & 129B & 224B & 137 ms\\
RetinaNet(s1a1) & 31.0 & 98B & 193B & 99 ms\\ 
Pruned & 33.9& 31B & 126B & 79 ms \\
Pruned(+\textit{retrained}) & 35.1& \textbf{31B} & 126B & 79 ms \\ \bottomrule
\end{tabular}
\label{tab:retinanet}
\end{table}
Table \ref{tab:retinanet} compares the result of retraining this FLOPs frontier configuration to the baseline and a RetinaNet version with only 1 anchor in each layer as reported in the original paper. Not only does our pruned configuration achieve 4.1\% better accuracy, it can also reduce the computational cost of the head by 75\% compared to the RetinaNet (s3a3) baseline and by 68\% compared to RetinaNet (s1a1).
This means that our method is able to\textbf{ reduce the FLOPs of the entire RetinaNet model by 44\% while only losing 1.8\% accuracy which is $3 \times$ better compression and 4.1\% better accuracy than the default way of scaling anchors}, \ie RetinaNet(s1a1), which only reduces the FLOPs by 14\% while also decreasing the accuracy by 5.9\%.

These results also illustrate the importance of adapting the anchor pruning search to the object detection architecture; without allowing an entire layer to be dropped it is impossible to significantly decrease the number of FLOPs.

The training configurations used for RetinaNet are based on the MMDetection \cite{mmdetection} RetinaNet baseline configuration with ResNet50 backbone and \texttt{$1\times$} learning rate schedule. All models are trained using SGD for 12 epochs with an initial learning rate of $0.01$ which is decreased to $10^{-3}$ on epoch 8 and $10^{-4}$ on epoch 11, the weight decay is $10^{-4}$ and the momentum $0.9$. The input images are resized to 800 pixels on the shortest side, and random horizontal flipping is used as data augmentation.

\subsection{MobileNetV2-SSDLite}
\begin{figure}[ht]
    \centering
    \includegraphics[width=.95\linewidth]{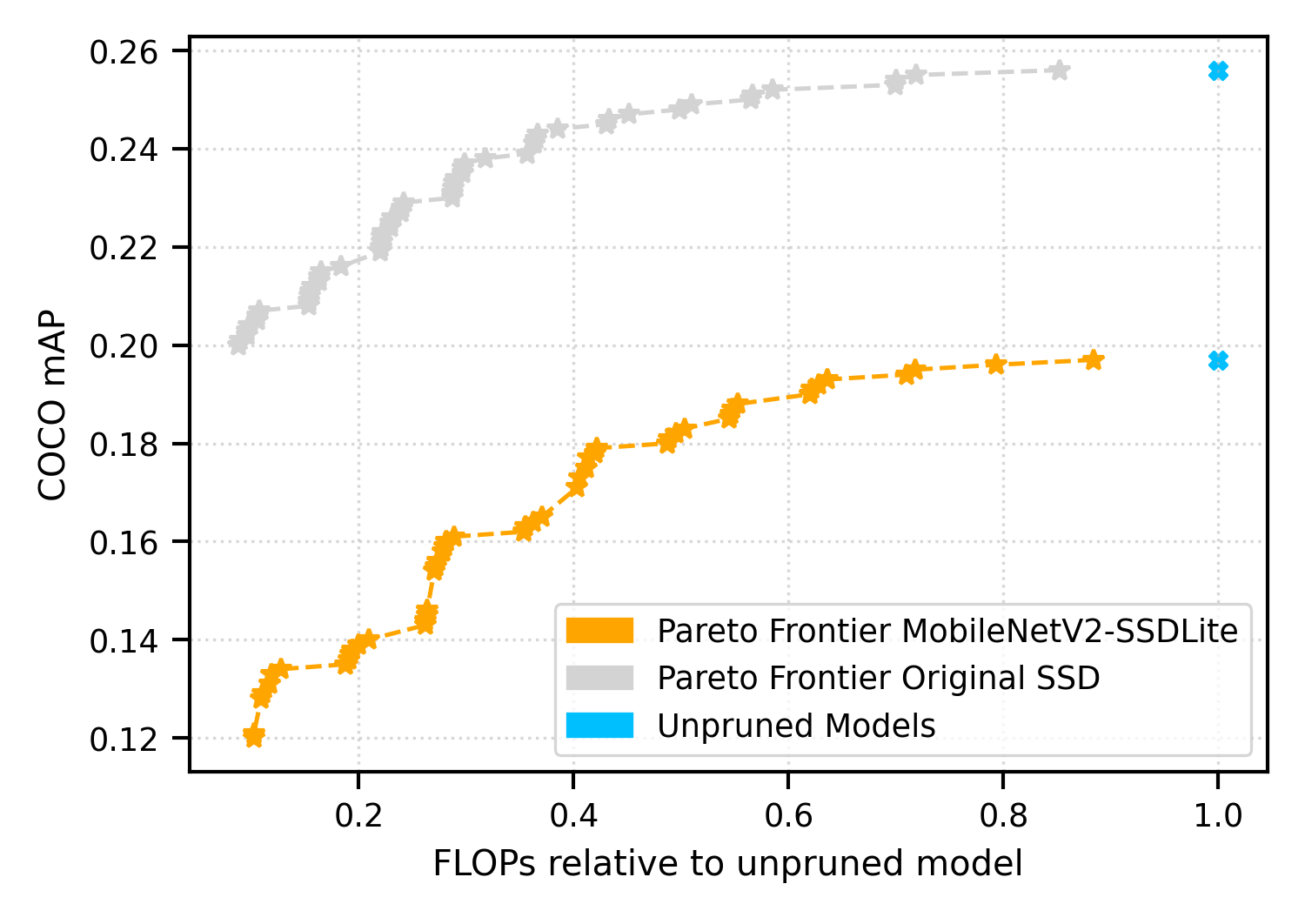}
    \caption{Accuracy and number of FLOPs relative to the unpruned models for MobileNetV2-SSDLite and the original SSD on COCO \texttt{val2017}. Both rontiers have a similar trend line, indicating that anchor pruning is backbone agnostic.}
    \label{fig:pareto_front_mobilenet}
\end{figure}

To further demonstrate the generalization of our approach we also conducted experiments on an SSD object detector with a more compact backbone. It is generally assumed that compact models such as MobileNet (\cite{sandler2018mobilenetv2}) and ShuffleNet (\cite{hu2018squeeze}) are harder to prune due to their already compact layers (\cite{zhu2017prune}).
For anchor pruning we show that the used backbone is not of importance as the pruning happens in the object detection head. The only important factor in anchor pruning is the placement of the original anchors. 
Figure \ref{fig:pareto_front_mobilenet} shows the Pareto Frontiers for MobileNetV2-SSDLite and the original SSD model. Both frontiers follow a similar trend line, indicating that anchor pruning is backbone agnostic. The main difference between the two trend lines is the slightly worse accuracy degradation on the left side in the MobileNet frontier. This is however explained by the fact that MobileNetV2-SSDLite model has only 3 anchors in the first layer compared to 4 in the original SSD model.

\subsection{PASCAL VOC}
For completeness and to further demonstrate the generalization of our approach, we also evaluate our approach on the PASCAL VOC (\cite{everingham2010pascal}) dataset. As is common, we use the combined datasets of VOC2007 and VOC2012.
To reproduce the SSD baseline in the same way as the original paper, we change the anchor scales to the values specific for PASCAL VOC as defined in (\cite{liu2016ssd}). The training configuration remains identical to the one used for COCO, with the exception that it is run for twice as many epochs.

\begin{figure}[ht]
    \centering
    \includegraphics[width=\linewidth]{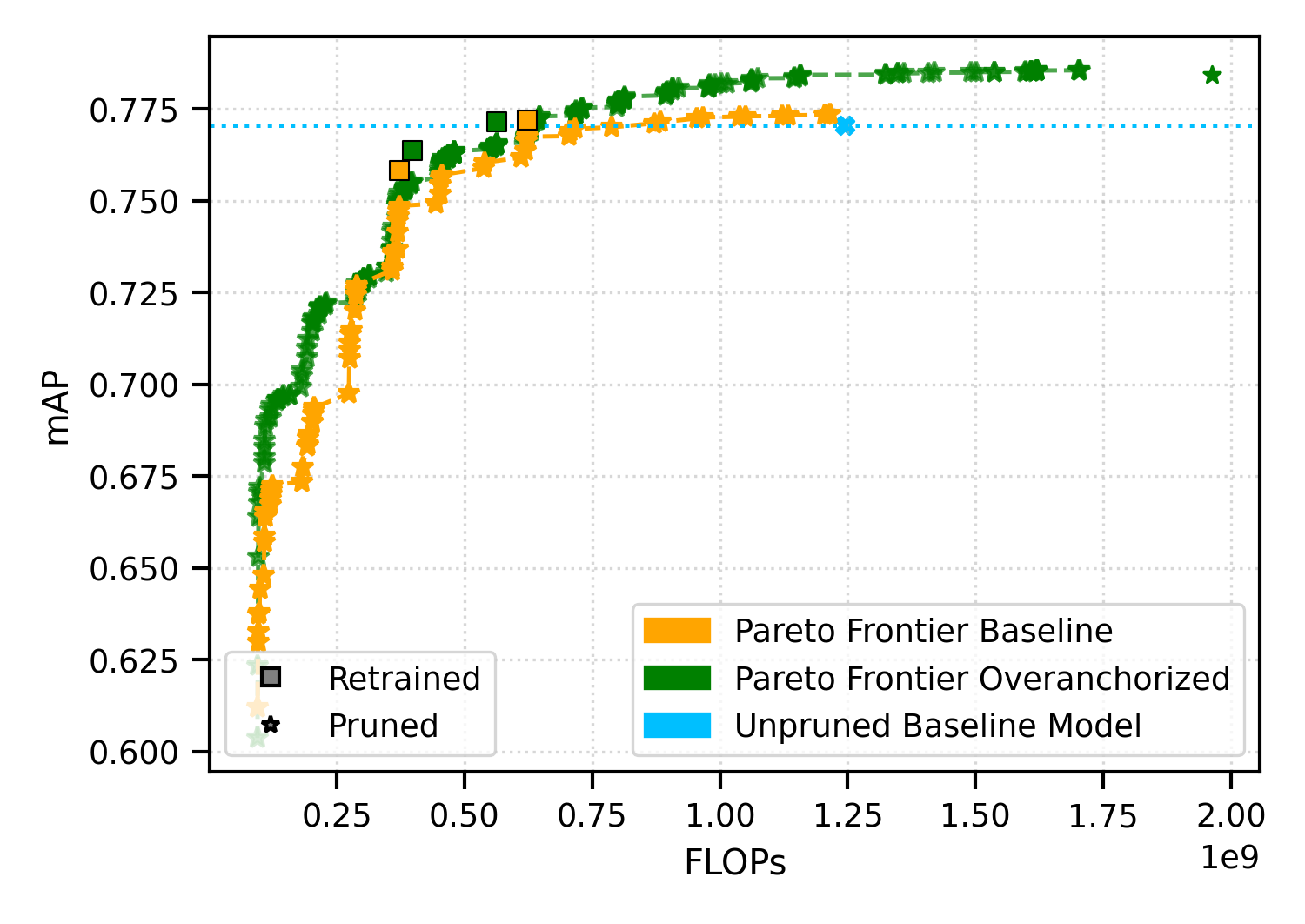}
    \caption{Accuracy and number of FLOPs for anchor pruning configurations in the Pareto frontiers for the SSD baseline and `overanchorized` model on PASCAL VOC. Note how pruning certain anchors results in improved accuracy compared to the baseline, even before any fine-tuning or retraining.}
    \label{fig:pareto_front_4}
\end{figure}

The Pareto frontier for SSD on this dataset can be found in Figure \ref{fig:pareto_front_4}. The reported mAP is with an IoU threshold of 0.50 as is custom for evaluating on this dataset. It can be seen that anchor pruning with retraining can \textbf{achieve similar mAP as the baseline with more than 50\% FLOPs reduction}.

As SSD uses different anchor shape initialization for PASCAL VOC compared to COCO, we also included an `overanchorized' model that has the same initial anchors as in the experiments of Section \ref{sec:overanchorized}. For both PASCAL VOC and COCO pruning from the same `overanchorized' model results in an improvement over the baseline models. This demonstrates that anchor pruning can not only eliminate the number of anchors as parameters but also the shapes of the anchors when used with an `overanchorized' detection head.

\begin{figure}[ht]
    \centering
    \subfigure[]{\includegraphics[width=.45\linewidth]{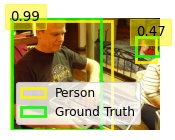}}
    \subfigure[]{\includegraphics[width=.45\linewidth]{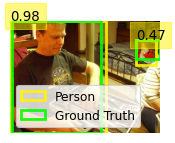}}
    \caption{An example illustrating how predictions made by a large anchor can be more confident but less accurate (a) than predictions made by a smaller anchor in an earlier layer (b).}
    \label{fig:voc_examples}
\end{figure}

Compared to the results on COCO, pruning SSD on PASCAL VOC shows an additional interesting property; removing certain anchors immediately improves accuracy, even before fine-tuning or retraining. This can be explained by looking at the pruned anchors. For example, the most accurate configuration prunes, among others, all anchors in the last layer. In SSD the last layer of the head operates on a $1 \times 1$ feature map while the previous layer operates on a $3 \times 3$ feature map. 
When anchors from both these layers make a prediction about the same large object, the anchor from the layer with the larger feature map has more spatial information which can lead to more accurate bounding box predictions.
The example image in Figure \ref{fig:voc_examples} illustrates this: while the anchor from the last layer is more confident in its prediction, the anchor from the previous layer is more accurate in predicting the bounding box shape. 
Depending on the network structure in the backbone and neck, removing all anchors on the last feature map may make this layer obsolete and reduce the computational complexity of the entire network even further.

\section{Conclusion}
In this paper, we proposed a novel pruning method for one-stage anchor-based object detection models: anchor pruning. 
We show that in most object detectors many anchors are redundant and that pruning those anchors followed by a fine-tuning or retraining step can increase accuracy while simultaneously reducing the computational cost.
Through a simple yet effective search algorithm, we provide a Pareto frontier of anchor configurations, allowing model designers to trade accuracy for performance when resources are constrained. We demonstrate the effects of pruning anchors extensively on the SSD and RetinaNet object detectors and the MS COCO and PASCAL VOC datasets. We also show that through pruning an `overanchorized' model we avoid tuning hyperparameters related to the initial shapes of the anchors. Given its effectiveness to make object detection models more efficient, we hope that anchor pruning will become part of the design process for modern one-stage object detectors.

\section*{Acknowledgments}
This research received funding through the imec.icon project cREAtIve and the Research Foundation Flanders (FWO-Vlaanderen) under Grant G006718N and 1S47820N.

\bibliographystyle{model2-names}
\bibliography{refs}

\begin{thebibliography}{39}
\expandafter\ifx\csname natexlab\endcsname\relax\def\natexlab#1{#1}\fi
\providecommand{\url}[1]{\texttt{#1}}
\providecommand{\href}[2]{#2}
\providecommand{\path}[1]{#1}
\providecommand{\DOIprefix}{doi:}
\providecommand{\ArXivprefix}{arXiv:}
\providecommand{\URLprefix}{URL: }
\providecommand{\Pubmedprefix}{pmid:}
\providecommand{\doi}[1]{\href{http://dx.doi.org/#1}{\path{#1}}}
\providecommand{\Pubmed}[1]{\href{pmid:#1}{\path{#1}}}
\providecommand{\bibinfo}[2]{#2}
\ifx\xfnm\relax \def\xfnm[#1]{\unskip,\space#1}\fi
\bibitem[{Cai et~al.(2019)Cai, Zhao, Wang, Lin, Foo, Aly and
  Chandrasekhar}]{cai2019maxpoolnms}
\bibinfo{author}{Cai, L.}, \bibinfo{author}{Zhao, B.}, \bibinfo{author}{Wang,
  Z.}, \bibinfo{author}{Lin, J.}, \bibinfo{author}{Foo, C.S.},
  \bibinfo{author}{Aly, M.S.}, \bibinfo{author}{Chandrasekhar, V.},
  \bibinfo{year}{2019}.
\newblock \bibinfo{title}{Maxpoolnms: getting rid of nms bottlenecks in
  two-stage object detectors}, in: \bibinfo{booktitle}{Proceedings of the IEEE
  Conference on Computer Vision and Pattern Recognition}, pp.
  \bibinfo{pages}{9356--9364}.
\bibitem[{Cai and Vasconcelos(2018)}]{cai2018cascade}
\bibinfo{author}{Cai, Z.}, \bibinfo{author}{Vasconcelos, N.},
  \bibinfo{year}{2018}.
\newblock \bibinfo{title}{Cascade r-cnn: Delving into high quality object
  detection}, in: \bibinfo{booktitle}{Proceedings of the IEEE conference on
  computer vision and pattern recognition}, pp. \bibinfo{pages}{6154--6162}.
\bibitem[{Chen et~al.(2019)Chen, Wang, Pang, Cao, Xiong, Li, Sun, Feng, Liu,
  Xu, Zhang, Cheng, Zhu, Cheng, Zhao, Li, Lu, Zhu, Wu, Dai, Wang, Shi, Ouyang,
  Loy and Lin}]{mmdetection}
\bibinfo{author}{Chen, K.}, \bibinfo{author}{Wang, J.}, \bibinfo{author}{Pang,
  J.}, \bibinfo{author}{Cao, Y.}, \bibinfo{author}{Xiong, Y.},
  \bibinfo{author}{Li, X.}, \bibinfo{author}{Sun, S.}, \bibinfo{author}{Feng,
  W.}, \bibinfo{author}{Liu, Z.}, \bibinfo{author}{Xu, J.},
  \bibinfo{author}{Zhang, Z.}, \bibinfo{author}{Cheng, D.},
  \bibinfo{author}{Zhu, C.}, \bibinfo{author}{Cheng, T.},
  \bibinfo{author}{Zhao, Q.}, \bibinfo{author}{Li, B.}, \bibinfo{author}{Lu,
  X.}, \bibinfo{author}{Zhu, R.}, \bibinfo{author}{Wu, Y.},
  \bibinfo{author}{Dai, J.}, \bibinfo{author}{Wang, J.}, \bibinfo{author}{Shi,
  J.}, \bibinfo{author}{Ouyang, W.}, \bibinfo{author}{Loy, C.C.},
  \bibinfo{author}{Lin, D.}, \bibinfo{year}{2019}.
\newblock \bibinfo{title}{{MMDetection}: Open mmlab detection toolbox and
  benchmark}.
\newblock \bibinfo{journal}{arXiv preprint arXiv:1906.07155} .
\bibitem[{Deng et~al.(2020)Deng, Li, Han, Shi and Xie}]{deng2020model}
\bibinfo{author}{Deng, L.}, \bibinfo{author}{Li, G.}, \bibinfo{author}{Han,
  S.}, \bibinfo{author}{Shi, L.}, \bibinfo{author}{Xie, Y.},
  \bibinfo{year}{2020}.
\newblock \bibinfo{title}{Model compression and hardware acceleration for
  neural networks: A comprehensive survey}.
\newblock \bibinfo{journal}{Proceedings of the IEEE} \bibinfo{volume}{108},
  \bibinfo{pages}{485--532}.
\bibitem[{Everingham et~al.(2010)Everingham, Van~Gool, Williams, Winn and
  Zisserman}]{everingham2010pascal}
\bibinfo{author}{Everingham, M.}, \bibinfo{author}{Van~Gool, L.},
  \bibinfo{author}{Williams, C.K.}, \bibinfo{author}{Winn, J.},
  \bibinfo{author}{Zisserman, A.}, \bibinfo{year}{2010}.
\newblock \bibinfo{title}{The pascal visual object classes (voc) challenge}.
\newblock \bibinfo{journal}{International journal of computer vision}
  \bibinfo{volume}{88}, \bibinfo{pages}{303--338}.
\bibitem[{Girshick(2015)}]{girshick2015fast}
\bibinfo{author}{Girshick, R.}, \bibinfo{year}{2015}.
\newblock \bibinfo{title}{Fast r-cnn}, in: \bibinfo{booktitle}{Proceedings of
  the IEEE international conference on computer vision}, pp.
  \bibinfo{pages}{1440--1448}.
\bibitem[{He et~al.(2016)He, Zhang, Ren and Sun}]{he2016deep}
\bibinfo{author}{He, K.}, \bibinfo{author}{Zhang, X.}, \bibinfo{author}{Ren,
  S.}, \bibinfo{author}{Sun, J.}, \bibinfo{year}{2016}.
\newblock \bibinfo{title}{Deep residual learning for image recognition}, in:
  \bibinfo{booktitle}{Proceedings of the IEEE conference on computer vision and
  pattern recognition}, pp. \bibinfo{pages}{770--778}.
\bibitem[{Hinton et~al.(2015)Hinton, Vinyals and Dean}]{hinton2015distilling}
\bibinfo{author}{Hinton, G.}, \bibinfo{author}{Vinyals, O.},
  \bibinfo{author}{Dean, J.}, \bibinfo{year}{2015}.
\newblock \bibinfo{title}{Distilling the knowledge in a neural network}.
\newblock \bibinfo{journal}{arXiv preprint arXiv:1503.02531} .
\bibitem[{Hu et~al.(2018)Hu, Shen and Sun}]{hu2018squeeze}
\bibinfo{author}{Hu, J.}, \bibinfo{author}{Shen, L.}, \bibinfo{author}{Sun,
  G.}, \bibinfo{year}{2018}.
\newblock \bibinfo{title}{Squeeze-and-excitation networks}, in:
  \bibinfo{booktitle}{Proceedings of the IEEE conference on computer vision and
  pattern recognition}, pp. \bibinfo{pages}{7132--7141}.
\bibitem[{Huang et~al.(2017)Huang, Rathod, Sun, Zhu, Korattikara, Fathi,
  Fischer, Wojna, Song, Guadarrama et~al.}]{huang2017speed}
\bibinfo{author}{Huang, J.}, \bibinfo{author}{Rathod, V.},
  \bibinfo{author}{Sun, C.}, \bibinfo{author}{Zhu, M.},
  \bibinfo{author}{Korattikara, A.}, \bibinfo{author}{Fathi, A.},
  \bibinfo{author}{Fischer, I.}, \bibinfo{author}{Wojna, Z.},
  \bibinfo{author}{Song, Y.}, \bibinfo{author}{Guadarrama, S.}, et~al.,
  \bibinfo{year}{2017}.
\newblock \bibinfo{title}{Speed/accuracy trade-offs for modern convolutional
  object detectors}, in: \bibinfo{booktitle}{Proceedings of the IEEE conference
  on computer vision and pattern recognition}, pp. \bibinfo{pages}{7310--7311}.
\bibitem[{Ke et~al.(2020)Ke, Zhang, Huang, Ye, Liu and Huang}]{ke2020multiple}
\bibinfo{author}{Ke, W.}, \bibinfo{author}{Zhang, T.}, \bibinfo{author}{Huang,
  Z.}, \bibinfo{author}{Ye, Q.}, \bibinfo{author}{Liu, J.},
  \bibinfo{author}{Huang, D.}, \bibinfo{year}{2020}.
\newblock \bibinfo{title}{Multiple anchor learning for visual object
  detection}, in: \bibinfo{booktitle}{Proceedings of the IEEE/CVF Conference on
  Computer Vision and Pattern Recognition}, pp. \bibinfo{pages}{10206--10215}.
\bibitem[{Law and Deng(2018)}]{law2018cornernet}
\bibinfo{author}{Law, H.}, \bibinfo{author}{Deng, J.}, \bibinfo{year}{2018}.
\newblock \bibinfo{title}{Cornernet: Detecting objects as paired keypoints},
  in: \bibinfo{booktitle}{Proceedings of the European Conference on Computer
  Vision (ECCV)}, pp. \bibinfo{pages}{734--750}.
\bibitem[{Li et~al.(2016)Li, Kadav, Durdanovic, Samet and Graf}]{li2016pruning}
\bibinfo{author}{Li, H.}, \bibinfo{author}{Kadav, A.},
  \bibinfo{author}{Durdanovic, I.}, \bibinfo{author}{Samet, H.},
  \bibinfo{author}{Graf, H.P.}, \bibinfo{year}{2016}.
\newblock \bibinfo{title}{Pruning filters for efficient convnets}.
\newblock \bibinfo{journal}{arXiv preprint arXiv:1608.08710} .
\bibitem[{Li et~al.(2019)Li, Yang, Huang, Hua and Zhang}]{li2019dynamic}
\bibinfo{author}{Li, S.}, \bibinfo{author}{Yang, L.}, \bibinfo{author}{Huang,
  J.}, \bibinfo{author}{Hua, X.S.}, \bibinfo{author}{Zhang, L.},
  \bibinfo{year}{2019}.
\newblock \bibinfo{title}{Dynamic anchor feature selection for single-shot
  object detection}, in: \bibinfo{booktitle}{Proceedings of the IEEE
  International Conference on Computer Vision}, pp.
  \bibinfo{pages}{6609--6618}.
\bibitem[{Liao et~al.(2018)Liao, Shi and Bai}]{liao2018textboxes++}
\bibinfo{author}{Liao, M.}, \bibinfo{author}{Shi, B.}, \bibinfo{author}{Bai,
  X.}, \bibinfo{year}{2018}.
\newblock \bibinfo{title}{Textboxes++: A single-shot oriented scene text
  detector}.
\newblock \bibinfo{journal}{IEEE transactions on image processing}
  \bibinfo{volume}{27}, \bibinfo{pages}{3676--3690}.
\bibitem[{Lin et~al.(2017a)Lin, Doll{\'a}r, Girshick, He, Hariharan and
  Belongie}]{lin2017feature}
\bibinfo{author}{Lin, T.Y.}, \bibinfo{author}{Doll{\'a}r, P.},
  \bibinfo{author}{Girshick, R.}, \bibinfo{author}{He, K.},
  \bibinfo{author}{Hariharan, B.}, \bibinfo{author}{Belongie, S.},
  \bibinfo{year}{2017}a.
\newblock \bibinfo{title}{Feature pyramid networks for object detection}, in:
  \bibinfo{booktitle}{Proceedings of the IEEE conference on computer vision and
  pattern recognition}, pp. \bibinfo{pages}{2117--2125}.
\bibitem[{Lin et~al.(2017b)Lin, Goyal, Girshick, He and
  Doll{\'a}r}]{lin2017retinanet}
\bibinfo{author}{Lin, T.Y.}, \bibinfo{author}{Goyal, P.},
  \bibinfo{author}{Girshick, R.}, \bibinfo{author}{He, K.},
  \bibinfo{author}{Doll{\'a}r, P.}, \bibinfo{year}{2017}b.
\newblock \bibinfo{title}{Focal loss for dense object detection}, in:
  \bibinfo{booktitle}{Proceedings of the IEEE international conference on
  computer vision}, pp. \bibinfo{pages}{2980--2988}.
\bibitem[{Lin et~al.(2014)Lin, Maire, Belongie, Hays, Perona, Ramanan,
  Doll{\'a}r and Zitnick}]{lin2014microsoft}
\bibinfo{author}{Lin, T.Y.}, \bibinfo{author}{Maire, M.},
  \bibinfo{author}{Belongie, S.}, \bibinfo{author}{Hays, J.},
  \bibinfo{author}{Perona, P.}, \bibinfo{author}{Ramanan, D.},
  \bibinfo{author}{Doll{\'a}r, P.}, \bibinfo{author}{Zitnick, C.L.},
  \bibinfo{year}{2014}.
\newblock \bibinfo{title}{Microsoft coco: Common objects in context}, in:
  \bibinfo{booktitle}{European conference on computer vision},
  \bibinfo{organization}{Springer}. pp. \bibinfo{pages}{740--755}.
\bibitem[{Liu et~al.(2016)Liu, Anguelov, Erhan, Szegedy, Reed, Fu and
  Berg}]{liu2016ssd}
\bibinfo{author}{Liu, W.}, \bibinfo{author}{Anguelov, D.},
  \bibinfo{author}{Erhan, D.}, \bibinfo{author}{Szegedy, C.},
  \bibinfo{author}{Reed, S.}, \bibinfo{author}{Fu, C.Y.},
  \bibinfo{author}{Berg, A.C.}, \bibinfo{year}{2016}.
\newblock \bibinfo{title}{Ssd: Single shot multibox detector}, in:
  \bibinfo{booktitle}{European conference on computer vision},
  \bibinfo{organization}{Springer}. pp. \bibinfo{pages}{21--37}.
\bibitem[{Liu et~al.(2018)Liu, Sun, Zhou, Huang and
  Darrell}]{liu2018rethinking}
\bibinfo{author}{Liu, Z.}, \bibinfo{author}{Sun, M.}, \bibinfo{author}{Zhou,
  T.}, \bibinfo{author}{Huang, G.}, \bibinfo{author}{Darrell, T.},
  \bibinfo{year}{2018}.
\newblock \bibinfo{title}{Rethinking the value of network pruning}.
\newblock \bibinfo{journal}{arXiv preprint arXiv:1810.05270} .
\bibitem[{Luo et~al.(2016)Luo, Li, Urtasun and Zemel}]{luo2016understanding}
\bibinfo{author}{Luo, W.}, \bibinfo{author}{Li, Y.}, \bibinfo{author}{Urtasun,
  R.}, \bibinfo{author}{Zemel, R.}, \bibinfo{year}{2016}.
\newblock \bibinfo{title}{Understanding the effective receptive field in deep
  convolutional neural networks}, in: \bibinfo{booktitle}{Advances in neural
  information processing systems}, pp. \bibinfo{pages}{4898--4906}.
\bibitem[{Redmon et~al.(2016)Redmon, Divvala, Girshick and
  Farhadi}]{redmon2016you}
\bibinfo{author}{Redmon, J.}, \bibinfo{author}{Divvala, S.},
  \bibinfo{author}{Girshick, R.}, \bibinfo{author}{Farhadi, A.},
  \bibinfo{year}{2016}.
\newblock \bibinfo{title}{You only look once: Unified, real-time object
  detection}, in: \bibinfo{booktitle}{Proceedings of the IEEE conference on
  computer vision and pattern recognition}, pp. \bibinfo{pages}{779--788}.
\bibitem[{Redmon and Farhadi(2017)}]{redmon2017yolo9000}
\bibinfo{author}{Redmon, J.}, \bibinfo{author}{Farhadi, A.},
  \bibinfo{year}{2017}.
\newblock \bibinfo{title}{Yolo9000: better, faster, stronger}, in:
  \bibinfo{booktitle}{Proceedings of the IEEE conference on computer vision and
  pattern recognition}, pp. \bibinfo{pages}{7263--7271}.
\bibitem[{Redmon and Farhadi(2018)}]{redmon2018yolov3}
\bibinfo{author}{Redmon, J.}, \bibinfo{author}{Farhadi, A.},
  \bibinfo{year}{2018}.
\newblock \bibinfo{title}{Yolov3: An incremental improvement}.
\newblock \bibinfo{journal}{arXiv preprint arXiv:1804.02767} .
\bibitem[{Ren et~al.(2015)Ren, He, Girshick and Sun}]{ren2015faster}
\bibinfo{author}{Ren, S.}, \bibinfo{author}{He, K.}, \bibinfo{author}{Girshick,
  R.}, \bibinfo{author}{Sun, J.}, \bibinfo{year}{2015}.
\newblock \bibinfo{title}{Faster r-cnn: Towards real-time object detection with
  region proposal networks}, in: \bibinfo{booktitle}{Advances in neural
  information processing systems}, pp. \bibinfo{pages}{91--99}.
\bibitem[{Sandler et~al.(2018)Sandler, Howard, Zhu, Zhmoginov and
  Chen}]{sandler2018mobilenetv2}
\bibinfo{author}{Sandler, M.}, \bibinfo{author}{Howard, A.},
  \bibinfo{author}{Zhu, M.}, \bibinfo{author}{Zhmoginov, A.},
  \bibinfo{author}{Chen, L.C.}, \bibinfo{year}{2018}.
\newblock \bibinfo{title}{Mobilenetv2: Inverted residuals and linear
  bottlenecks}, in: \bibinfo{booktitle}{Proceedings of the IEEE conference on
  computer vision and pattern recognition}, pp. \bibinfo{pages}{4510--4520}.
\bibitem[{Sermanet et~al.(2013)Sermanet, Eigen, Zhang, Mathieu, Fergus and
  LeCun}]{sermanet2013overfeat}
\bibinfo{author}{Sermanet, P.}, \bibinfo{author}{Eigen, D.},
  \bibinfo{author}{Zhang, X.}, \bibinfo{author}{Mathieu, M.},
  \bibinfo{author}{Fergus, R.}, \bibinfo{author}{LeCun, Y.},
  \bibinfo{year}{2013}.
\newblock \bibinfo{title}{Overfeat: Integrated recognition, localization and
  detection using convolutional networks}.
\newblock \bibinfo{journal}{arXiv preprint arXiv:1312.6229} .
\bibitem[{Simonyan and Zisserman(2014)}]{simonyan2014very}
\bibinfo{author}{Simonyan, K.}, \bibinfo{author}{Zisserman, A.},
  \bibinfo{year}{2014}.
\newblock \bibinfo{title}{Very deep convolutional networks for large-scale
  image recognition}.
\newblock \bibinfo{journal}{arXiv preprint arXiv:1409.1556} .
\bibitem[{Tan et~al.(2020)Tan, Pang and Le}]{tan2020efficientdet}
\bibinfo{author}{Tan, M.}, \bibinfo{author}{Pang, R.}, \bibinfo{author}{Le,
  Q.V.}, \bibinfo{year}{2020}.
\newblock \bibinfo{title}{Efficientdet: Scalable and efficient object
  detection}, in: \bibinfo{booktitle}{Proceedings of the IEEE/CVF Conference on
  Computer Vision and Pattern Recognition}, pp. \bibinfo{pages}{10781--10790}.
\bibitem[{Tian et~al.(2019)Tian, Shen, Chen and He}]{tian2019fcos}
\bibinfo{author}{Tian, Z.}, \bibinfo{author}{Shen, C.}, \bibinfo{author}{Chen,
  H.}, \bibinfo{author}{He, T.}, \bibinfo{year}{2019}.
\newblock \bibinfo{title}{Fcos: Fully convolutional one-stage object
  detection}, in: \bibinfo{booktitle}{Proceedings of the IEEE international
  conference on computer vision}, pp. \bibinfo{pages}{9627--9636}.
\bibitem[{Verucchi et~al.(2020)Verucchi, Brilli, Sapienza, Verasani, Arena,
  Gatti, Capotondi, Cavicchioli, Bertogna and Solieri}]{verucchi2020systematic}
\bibinfo{author}{Verucchi, M.}, \bibinfo{author}{Brilli, G.},
  \bibinfo{author}{Sapienza, D.}, \bibinfo{author}{Verasani, M.},
  \bibinfo{author}{Arena, M.}, \bibinfo{author}{Gatti, F.},
  \bibinfo{author}{Capotondi, A.}, \bibinfo{author}{Cavicchioli, R.},
  \bibinfo{author}{Bertogna, M.}, \bibinfo{author}{Solieri, M.},
  \bibinfo{year}{2020}.
\newblock \bibinfo{title}{A systematic assessment of embedded neural networks
  for object detection}, in: \bibinfo{booktitle}{2020 25th IEEE International
  Conference on Emerging Technologies and Factory Automation (ETFA)},
  \bibinfo{organization}{IEEE}. pp. \bibinfo{pages}{937--944}.
\bibitem[{Wang et~al.(2019)Wang, Chen, Yang, Loy and
  Lin}]{wang2019guidedanchoring}
\bibinfo{author}{Wang, J.}, \bibinfo{author}{Chen, K.}, \bibinfo{author}{Yang,
  S.}, \bibinfo{author}{Loy, C.C.}, \bibinfo{author}{Lin, D.},
  \bibinfo{year}{2019}.
\newblock \bibinfo{title}{Region proposal by guided anchoring}, in:
  \bibinfo{booktitle}{Proceedings of the IEEE Conference on Computer Vision and
  Pattern Recognition}, pp. \bibinfo{pages}{2965--2974}.
\bibitem[{Yang et~al.(2018)Yang, Zhang, Li, Zhang and Sun}]{yang2018metaanchor}
\bibinfo{author}{Yang, T.}, \bibinfo{author}{Zhang, X.}, \bibinfo{author}{Li,
  Z.}, \bibinfo{author}{Zhang, W.}, \bibinfo{author}{Sun, J.},
  \bibinfo{year}{2018}.
\newblock \bibinfo{title}{Metaanchor: Learning to detect objects with
  customized anchors}, in: \bibinfo{booktitle}{Advances in Neural Information
  Processing Systems}, pp. \bibinfo{pages}{320--330}.
\bibitem[{Zhang et~al.(2018)Zhang, Wen, Bian, Lei and Li}]{zhang2018refinedet}
\bibinfo{author}{Zhang, S.}, \bibinfo{author}{Wen, L.}, \bibinfo{author}{Bian,
  X.}, \bibinfo{author}{Lei, Z.}, \bibinfo{author}{Li, S.Z.},
  \bibinfo{year}{2018}.
\newblock \bibinfo{title}{Single-shot refinement neural network for object
  detection}, in: \bibinfo{booktitle}{Proceedings of the IEEE conference on
  computer vision and pattern recognition}, pp. \bibinfo{pages}{4203--4212}.
\bibitem[{Zhang et~al.(2017)Zhang, Zhu, Lei, Shi, Wang and Li}]{zhang2017s3fd}
\bibinfo{author}{Zhang, S.}, \bibinfo{author}{Zhu, X.}, \bibinfo{author}{Lei,
  Z.}, \bibinfo{author}{Shi, H.}, \bibinfo{author}{Wang, X.},
  \bibinfo{author}{Li, S.Z.}, \bibinfo{year}{2017}.
\newblock \bibinfo{title}{S3fd: Single shot scale-invariant face detector}, in:
  \bibinfo{booktitle}{Proceedings of the IEEE international conference on
  computer vision}, pp. \bibinfo{pages}{192--201}.
\bibitem[{Zhang et~al.(2019)Zhang, Wan, Liu, Ji and Ye}]{zhang2019freeanchor}
\bibinfo{author}{Zhang, X.}, \bibinfo{author}{Wan, F.}, \bibinfo{author}{Liu,
  C.}, \bibinfo{author}{Ji, R.}, \bibinfo{author}{Ye, Q.},
  \bibinfo{year}{2019}.
\newblock \bibinfo{title}{Freeanchor: Learning to match anchors for visual
  object detection}, in: \bibinfo{booktitle}{Advances in Neural Information
  Processing Systems}, pp. \bibinfo{pages}{147--155}.
\bibitem[{Zhong et~al.(2020)Zhong, Wang, Peng and Zhang}]{zhong2020anchor}
\bibinfo{author}{Zhong, Y.}, \bibinfo{author}{Wang, J.}, \bibinfo{author}{Peng,
  J.}, \bibinfo{author}{Zhang, L.}, \bibinfo{year}{2020}.
\newblock \bibinfo{title}{Anchor box optimization for object detection}, in:
  \bibinfo{booktitle}{The IEEE Winter Conference on Applications of Computer
  Vision}, pp. \bibinfo{pages}{1286--1294}.
\bibitem[{Zhou et~al.(2017)Zhou, Yao, Guo, Xu and Chen}]{zhou2017incremental}
\bibinfo{author}{Zhou, A.}, \bibinfo{author}{Yao, A.}, \bibinfo{author}{Guo,
  Y.}, \bibinfo{author}{Xu, L.}, \bibinfo{author}{Chen, Y.},
  \bibinfo{year}{2017}.
\newblock \bibinfo{title}{Incremental network quantization: Towards lossless
  cnns with low-precision weights}.
\newblock \bibinfo{journal}{arXiv preprint arXiv:1702.03044} .
\bibitem[{Zhu and Gupta(2017)}]{zhu2017prune}
\bibinfo{author}{Zhu, M.}, \bibinfo{author}{Gupta, S.}, \bibinfo{year}{2017}.
\newblock \bibinfo{title}{To prune, or not to prune: exploring the efficacy of
  pruning for model compression}.
\newblock \bibinfo{journal}{arXiv preprint arXiv:1710.01878} .

\end{thebibliography}

\end{document}